\documentclass[11pt]{article}
\PassOptionsToPackage{shortlabels,inline}{enumitem}
\usepackage{agenticlearning-xelatex}
\hidelabreport
\setlist[itemize]{leftmargin=*}
\setlist[enumerate]{leftmargin=*}
\begin{document}
\shorttitle{Latent Merger}
\shortauthor{Yang \etal}
\title{Latent Merger: Latent-Space Composition of Hypernetwork-Generated LoRAs}
\author{
  Yanlai Yang$^{1}$, Minki Kang$^{2}$, Divyam Madaan$^{1}$,
  Sung Ju Hwang$^{2,3}$, and Mengye Ren$^{1}$ \\
  $^{1}$New York University, $^{2}$KAIST, $^{3}$DeepAuto.ai \\
  \texttt{yy2694@nyu.edu} \\
  \href{https://agenticlearning.ai/latent-merger}{\nolinkurl{agenticlearning.ai/latent-merger}}
}

\date{}\maketitle
{\let\thefootnote\relax\footnotetext{40th Conference on Neural Information Processing Systems (NeurIPS 2026).}}
\begin{abstract}
Recent work proposes hypernetworks that convert a document into a low-rank parameter update (LoRA) in a single forward pass, enabling fast injection of new knowledge into a frozen LLM. However, storing separate LoRA adapters for each document can be unsustainable in  streaming settings. In this paper, we study the problem of composing hypernetwork-generated LoRAs using two representative systems, Doc-to-LoRA and SHINE.
We show that hypernetwork-generated LoRAs have a geometry sharply different from directly fine-tuned LoRAs. They exhibit greater spectral concentration and stronger cross-document alignment than next-token prediction and QA fine-tuned adapters, which are diffuse and nearly orthogonal across documents. 
Guided by this observation, we propose latent-space LoRA merging. We compose the intermediate hypernetwork latents and then decode the merged latent through the frozen LoRA-generation head. We show that latent-space merging techniques consistently improve over weight-space and factor-space baselines while preserving a single fixed-rank adapter. These results support latent-space composition as a promising mechanism for combining hypernetwork-generated LoRAs under a fixed output-rank budget.
\end{abstract}

\section{Introduction}
Large language models (LLMs) have become increasingly capable at using long contexts, but incorporating new information into an LLM remains costly. 
A new document can be supplied in the prompt whenever it is needed, but this consumes context length, increases inference cost, and leaves the model's parameters unchanged. 
Alternatively, the document can be distilled into model parameters through fine-tuning~\citep{gururangan2020dont, hu2022lora, zweiger2025self, tang2026knowledge} or context distillation~\citep{snell2022learning}, but this often requires expensive per-document optimization. 
Hypernetwork-generated LoRAs~\citep{charakorn2026d2l,liu2026shine} offer a different interface for knowledge injection: after a one-time meta-training stage, a hypernetwork maps a document or context into a compact LoRA adapter in a single forward pass, while the base LLM remains frozen. The resulting adapter can then be stored, reused, and activated without placing the original document back into the prompt.

This interface is especially appealing for settings where knowledge arrives as a stream. 
For example, a deployed assistant may need to internalize a sequence of news articles, reports, user documents, or database records, and later answer many questions about this accumulated information. 
Generating one LoRA per document is a natural first step, but it creates a new composition problem. 
Maintaining an ever-growing bank of document adapters is inefficient, query-time routing to a single adapter is insufficient when questions may involve multiple documents, and activating many adapters by concatenation causes the effective rank, memory footprint, and update magnitude to grow with the number of documents. 
A practical document-to-LoRA system therefore needs a way to compose multiple generated adapters into a bounded representation.

Most existing model-merging methods operate directly in weight space or LoRA factor space, and are usually motivated by adapters obtained through independent gradient-based fine-tuning. 
Hypernetwork-generated LoRAs have a different origin. 
They are not arbitrary endpoints of separate optimization runs; they are outputs of a shared generator, often decoded from an intermediate latent representation. 

\begin{figure}[t]
  \centering
  \includegraphics[width=0.82\linewidth]{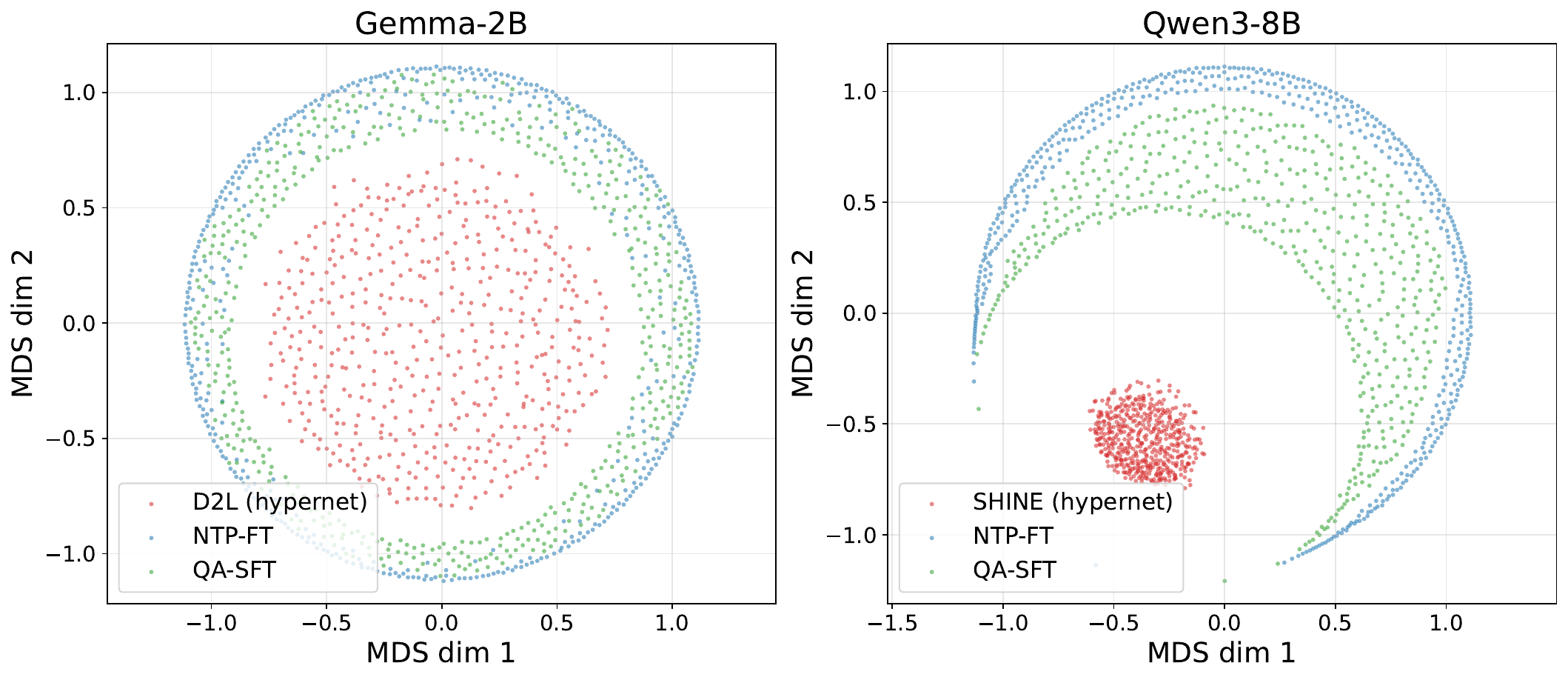}
  \caption{\textbf{Hypernetwork-generated LoRAs occupy a qualitatively different region of adapter space.}
  We visualize pairwise distances between unit-Frobenius-normalized adapters using metric MDS, with 500 passages per method sampled from the SQuAD validation set.
  In the two-dimensional MDS embedding, directly fine-tuned NTP-FT and QA-SFT adapters form broad shells, while D2L and SHINE adapters concentrate in compact regions.
  This structural difference motivates latent merging methods that respect the hypernetwork-induced adapter geometry.} 
  \label{fig:fig3b_mds_vis}
\end{figure}

In this work, we study this question using two recent hypernetwork-based LoRA systems for knowledge injection, Doc-to-LoRA (D2L)~\citep{charakorn2026d2l} and SHINE~\citep{liu2026shine}. 
We compare their document-specific adapters against LoRAs obtained by direct next-token prediction fine-tuning and QA supervised fine-tuning on the same passage set. 
Figure~\ref{fig:fig3b_mds_vis} shows a metric MDS embedding of pairwise adapter distances. 
In the two-dimensional MDS embedding, directly fine-tuned adapters form broad shells, while hypernetwork-generated adapters occupy more compact regions, especially for SHINE. 
This qualitative picture is confirmed by spectral and angular statistics in Section~\ref{sec:adapter_structure}: hypernetwork-generated adapters exhibit much stronger spectral concentration and substantially larger within-method cosine similarity than directly fine-tuned LoRAs.

Motivated by this observation, we propose Latent Merger, a method to merge hypernetwork-generated LoRAs in the latent space of the hypernetwork rather than in the decoded weight space. Given a set of documents, we extract their frozen hypernetwork latents, compose them with a learned, lightweight merging module, and then pass the merged latent through the frozen LoRA-generation head. Latent Merger is trained with the base LLM and the hypernetwork frozen. It produces a single fixed-rank LoRA adapter for the whole document set, rather than increasing rank or storing one adapter per document.

We evaluate this approach on multi-document QA tuples. 
Across both D2L and SHINE, Latent Merger consistently outperforms standard weight-space and factor-space merging methods. 
These results suggest that hypernetwork-generated LoRAs should be treated not merely as low-rank weight updates, but as decoded points on a structured adapter manifold whose latent coordinates provide a natural space for composition.

Our contributions are as follows:
\begin{itemize}
    \item We empirically characterize the geometry of hypernetwork-generated LoRAs, showing greater spectral concentration and cross-document alignment than next-token prediction and QA fine-tuned adapters. Metric MDS provides a qualitative visualization of these differences.
    \item We propose Latent Merger, a learned, lightweight latent-space LoRA merging module that composes document latents before decoding them through the frozen hypernetwork head.
    \item We show empirically that Latent Merger outperforms standard weight-space and factor-space merging baselines for multi-document QA while producing a single fixed-rank adapter.
\end{itemize}

\section{Related Work}

\paragraph{Hypernetwork-generated LoRAs.}
A growing line of work studies parameter-space updates that inject new knowledge into a pre-trained LLM. Earlier approaches include supervised fine-tuning on raw documents, document-specific question-answer pairs, or context distillation~\citep{askell2021general,mitchell2021fast,snell2022learning,mitchell2022memory,meng2022locating,meng2023massediting,qi2024context,shin2025generative}. These methods enable the model to answer queries without re-reading the original context, but they typically require expensive per-document optimization or synthetic data generation. Hypernetwork-generated LoRAs offer a complementary route by amortizing the update procedure: after meta-training, a single forward pass maps a document~\citep{charakorn2026d2l,liu2026shine}, task description~\citep{charakorn2025text}, or support set~\citep{phang2023hypertuning,ivison2023hint} to a compact adapter for a frozen backbone.

Most relevant to our work, Doc-to-LoRA (D2L) instantiates this idea for document internalization by meta-learning an approximate context-distillation procedure~\citep{charakorn2026d2l}. It encodes document-conditioned activations from a frozen LLM through a Perceiver-style bottleneck and decodes the resulting latent into LoRA factors, with a chunking mechanism that produces higher-rank adapters for long documents. SHINE~\citep{liu2026shine} pursues the same context-to-LoRA goal with a different architecture: it reuses the frozen LLM to extract memory states and applies a Transformer-based memory-to-parameter module whose structure mirrors the target LoRA parameters, enabling communication across layers and memory tokens before decoding the adapter. Our work studies how the document-specific adapters emitted by such hypernetworks can be merged. We use D2L and SHINE as architecturally distinct testbeds, but exploit their shared structure: both generate LoRA weights from an intermediate latent representation that can be more effectively composed than the LoRA weights themselves.

\paragraph{LoRA and model merging.}
Model merging seeks to combine independently fine-tuned models or adapters without joint retraining. Popular approaches include averaging checkpoints~\citep{wortsman2022model}, combining task vectors through task arithmetic~\citep{ilharco2022editing}, or using second-order~\citep{matena2022merging} or regression-based~\citep{jin2022dataless} criteria to form weighted averages. Some recent methods improve robustness by sparsifying updates, resolving sign conflicts, or dropping and rescaling delta weights~\citep{yadav2023ties,yu2024language}. These ideas have also been extended to the LoRA setting~\citep{huang2023lorahub,prabhakar2025lorasoups}. A complementary line of work exploits LoRA's low-rank structure more directly, for example by merging in singular-vector or singular-value space~\citep{gargiulo2025task,stoica2024model}, encouraging orthogonal adapter subspaces to reduce interference~\citep{buyukakyuz2024olora,zhang2025unraveling,xiong2026oplora,cao2026orthogonal}, searching for task-specific linear combination coefficients from a small validation set~\citep{huang2023lorahub,prabhakar2025lorasoups}. Some more recent methods also explore training a hypernetwork to predict the merge coefficients~\citep{shenaj2025lorarar} or the merged weights directly~\citep{zhang2025hyperadalora}. Also relevant to our work, KMerge~\citep{shenaj2025kmerge} considers budgeted sequential setting and maintains a bounded cache of adapters, merging similar LoRAs when capacity is reached; SLAO~\citep{qiao2025merge} continually folds new adapters into a single LoRA with orthogonal initialization and time-aware scaling. 

In contrast to these prior methods that operate in the weight or factor space, our method exploits the fact that each adapter is decoded by a frozen hypernetwork from a structured intermediate representation, and merges in this latent representation space directly. The merged latent code is then decoded through the frozen hypernetwork into the final LoRA. This latent-space formulation supports both training-free operators, such as averaging, and learned operators, such as a lightweight Transformer trained on document mixtures.

\paragraph{Composition in generative models.}
Related work in image generation studies operations on intermediate representations: StyleGAN mixes styles derived from learned latent codes~\citep{karras2019style}. Composable diffusion models combine conditional score estimates to compose visual concepts~\citep{liu2022compositional}. In this work, we present a lightweight learned composition rule for hypernetwork-generated LLM adapters under an output-rank budget.

\section{Structure of Hypernetwork-generated LoRA Adapters}
\label{sec:adapter_structure}

\begin{table}[t]
\centering
\caption{
Structural metrics for hypernetwork-generated and gradient-trained LoRAs. We report the average adapter Frobenius norm, the mean off-diagonal within-method cosine after unit-Frobenius normalization, and the fraction of uncentered spectral energy captured by the leading component.
}
\label{tab:structural_metrics}
\small
\begin{tabular}{llccc}
\toprule
Backbone & Method & $\|\Delta W\|_F$ & within cos. & top-1 energy \\
\midrule
\textit{Hypernetwork-based methods} & & & & \\
Gemma2-2B & D2L     & 30.2  & +0.144 & 15.5\% \\
Qwen3-8B & SHINE   & 209.4 & +0.888 & 88.8\% \\ \midrule
\textit{Gradient based methods} & & & & \\
Gemma2-2B & NTP-FT  & 5.6   & +0.005 & 0.6\%  \\
Qwen3-8B & NTP-FT  & 8.5   & +0.001 & 0.2\%  \\
Gemma2-2B & QA-SFT  & 3.5   & +0.014 & 1.5\%  \\
Qwen3-8B & QA-SFT  & 6.1   & +0.025 & 2.6\%  \\
\bottomrule
\end{tabular}
\end{table}

\begin{figure}[t]
    \centering
    \centering
    \includegraphics[width=0.82\linewidth]{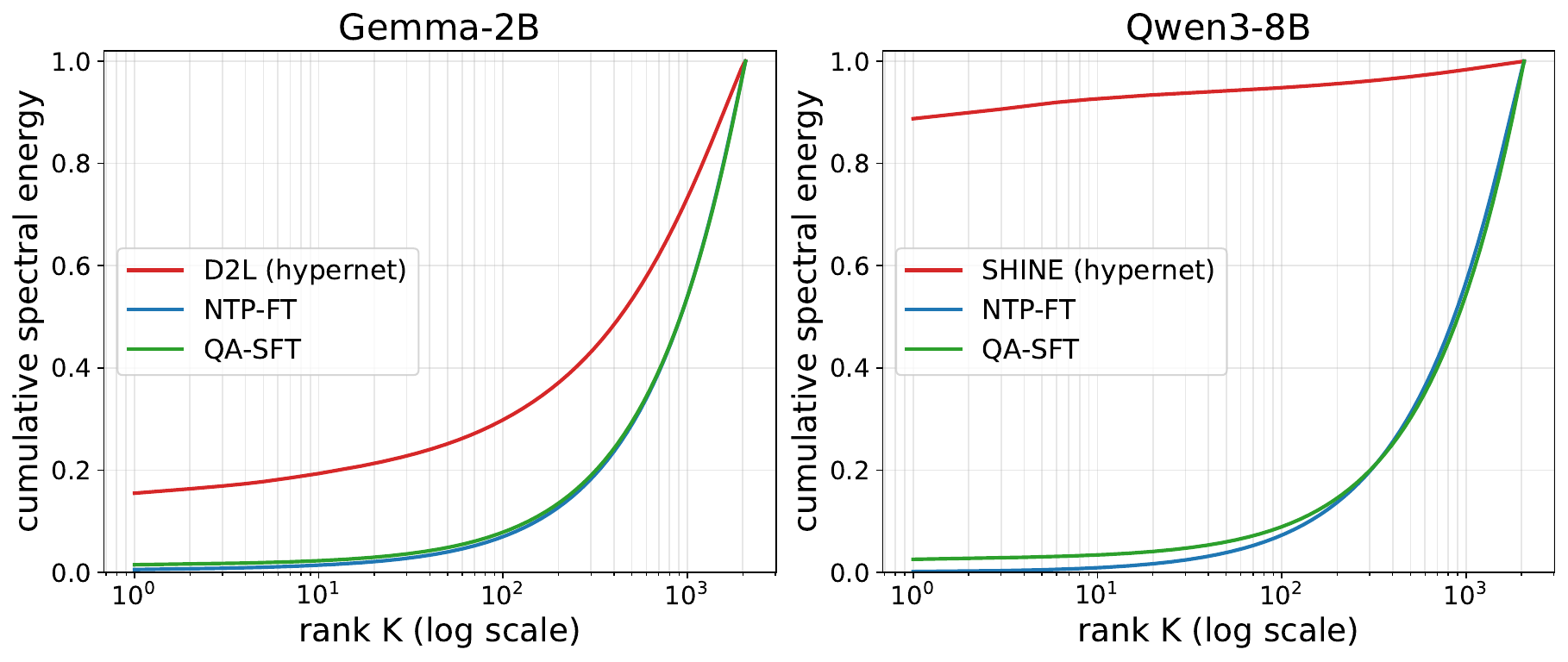}
    \caption{
    Cumulative spectral energy of stacked adapter matrices.
    Hypernetwork-generated adapters concentrate energy in fewer directions than directly fine-tuned NTP-FT and QA-SFT adapters.
    } 
    \label{fig:fig2_spectral_energy}
\end{figure}

Before studying how to merge multiple document adapters, we first examine whether hypernetwork-generated LoRAs have the same geometry as LoRAs obtained by direct fine-tuning. 
For each passage $p$ and method $m$, we concatenate the effective LoRA updates $\Delta W$ from all target layers $l$ and modules $a$ into a single vector
\[
    x_p^{(m)} = \mathrm{vec}\left(\{\Delta W_{p,\ell,a}^{(m)}\}_{\ell,a}\right).
\]

We analyze the resulting adapter bank through its Gram matrix $G_{ij}^{(m)} = \left\langle x_i^{(m)}, x_j^{(m)} \right\rangle$,
which we compute directly from the low-rank factors without materializing the dense updates. 
From this Gram matrix, we measure both the uncentered spectral energy,
\[
    E_m(K) =
    \frac{\sum_{k=1}^{K} \lambda_k(G^{(m)})}
    {\sum_{k} \lambda_k(G^{(m)})},
\]
and the angular geometry of unit-Frobenius-normalized adapters,
\[
    c_{ij}^{(m)}
    =
    \frac{G_{ij}^{(m)}}{\sqrt{G_{ii}^{(m)}G_{jj}^{(m)}}},
    \qquad
    d_{ij}^{(m)}
    =
    \left\|u_i^{(m)}-u_j^{(m)}\right\|_2
    =
    \sqrt{2(1-c_{ij}^{(m)})}.
\]
We compare D2L-generated adapters on Gemma2-2B~\citep{team2024gemma} and SHINE-generated adapters on Qwen3-8B~\citep{yang2025qwen3} against two direct fine-tuning controls on the same passage set: next-token prediction fine-tuning (NTP-FT) and QA supervised fine-tuning (QA-SFT).

Figure~\ref{fig:fig2_spectral_energy} shows that hypernetwork-generated adapters are substantially more spectrally concentrated than directly fine-tuned adapters. 
For both backbones, the spectral energy curve of the hypernetwork method lies far above the corresponding NTP-FT and QA-SFT curves, indicating that a small number of global directions explain a much larger fraction of the uncentered spectral energy. 
The effect is especially pronounced for SHINE on Qwen3-8B, whose spectrum is dominated by the leading component. 

The angular statistics reveal the same qualitative distinction. 
As shown in Table~\ref{tab:structural_metrics}, directly fine-tuned adapters have near-zero within-method cosine similarity, meaning that different passage-specific updates are almost orthogonal after unit-Frobenius normalization. 
By contrast, D2L adapters are more aligned with one another, and SHINE adapters are highly aligned. 
This shows that the spectral concentration above is not only a consequence of scale: even after normalizing each adapter to unit Frobenius norm, hypernetwork-generated adapters retain a much stronger shared direction than adapters obtained by direct fine-tuning. 
The Frobenius norms in Table~\ref{tab:structural_metrics} further show that the generated adapters also differ in magnitude from the fine-tuned controls, motivating our use of normalized cosine and distance metrics when comparing geometry.

Figure~\ref{fig:fig3b_mds_vis} provides a qualitative visualization of this angular geometry using metric MDS on pairwise unit-Frobenius distances. 
In the two-dimensional MDS embedding, the directly fine-tuned NTP-FT and QA-SFT adapters form broad shells, consistent with the near-zero cosine values in Table~\ref{tab:structural_metrics}. 
In contrast, the hypernetwork-generated adapters occupy more compact regions, with the SHINE adapters forming a particularly tight cluster. 

This structural difference has direct implications for adapter merging. 
If per-document LoRAs are nearly orthogonal, weight-space averaging attenuates each document-specific contribution, while concatenation retains the individual factors at a larger representation size. 
Hypernetwork-generated adapters exhibit stronger cross-document alignment and are decoded from latent representations by a shared generator. 
This shared parameterization motivates composing document representations before decoding them through the frozen hypernetwork head.

\section{Merging Hypernetwork-generated LoRAs}
\label{sec:merging}

We now study whether multiple document-specific hypernetwork LoRAs can be composed into a single adapter. Each evaluation example consists of a tuple of $N$ passages and their associated questions. A merger receives the adapters, or the corresponding hypernetwork latents, for all $N$ passages and must produce one LoRA dictionary shared across the whole tuple. At inference time, the model answers each question using this single merged adapter, without access to the original document in the context or the document-specific oracle adapter.

\subsection{Proposed Method: Learned Latent Merger}
\label{subsec:latent_merger}

\begin{figure}[t]
  \centering
  \includegraphics[width=\linewidth]{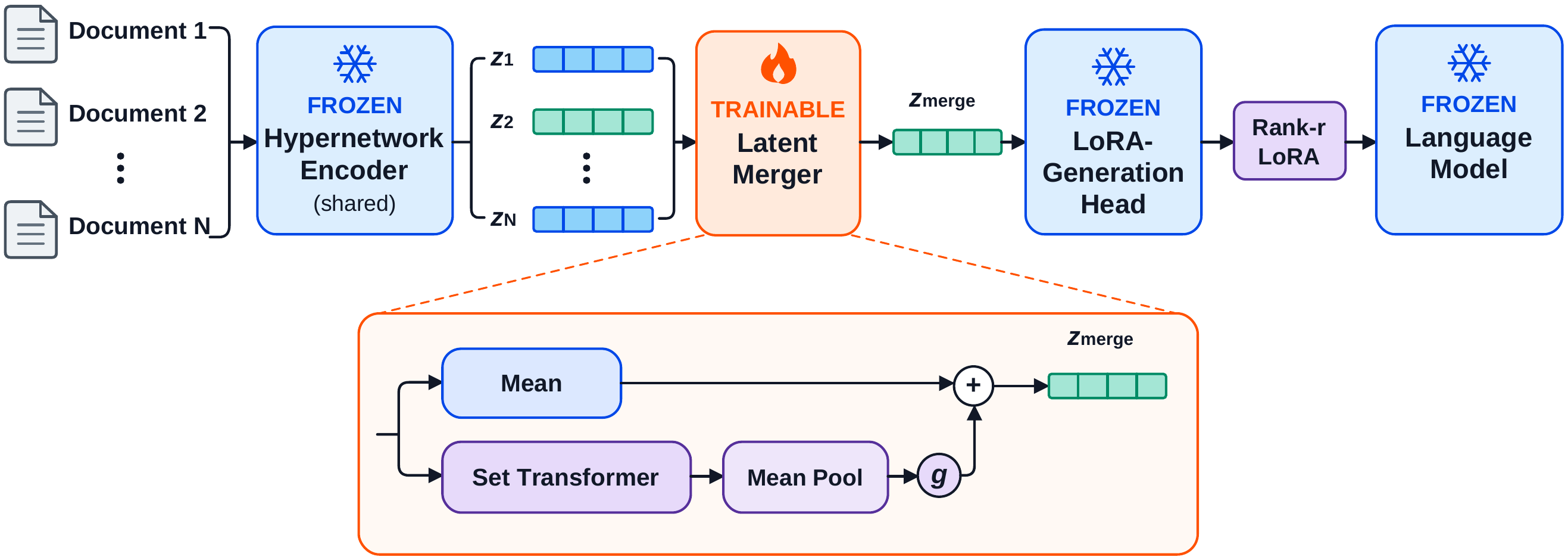}
  \caption{\textbf{Latent Merger overview.} A shared, frozen hypernetwork encoder maps each document to a latent. The trainable merger adds a gated, permutation-invariant residual to the latent mean. The frozen LoRA-generation head decodes the merged latent into one fixed-rank adapter for the frozen language model.}
  \label{fig:latent_merger_overview}
\end{figure}

Our hypothesis is that hypernetwork-generated LoRAs are easier to compose in the latent space than directly in the LoRA weight space. We therefore train Latent Merger, a lightweight cross-document merging module that composes multiple document-specific hypernetwork latents before the frozen LoRA-generation head. Figure~\ref{fig:latent_merger_overview} summarizes the approach.

Given a tuple of $N$ documents, we first extract the frozen hypernetwork latent for each document. We denote the latent at document slot $i$ and latent position $p$ by $z_{i,p} \in \mathbb{R}^{D}$, where $D$ is the latent feature dimension. The position index $p$ ranges over the non-feature axes of the corresponding hypernetwork latent: for D2L, one position corresponds to a single $(b,\ell,m,r)$ entry, while for SHINE, one position corresponds to a single $(\ell,t)$ entry. The merger is applied independently at each such latent position.

Our merger predicts a residual correction to the mean latent $m_p = \frac{1}{N}\sum_{i=1}^{N} z_{i,p}$. For each latent position $p$, we treat the $N$ document slots as a set. We first project each $z_{i,p}$ into a low-dimensional bottleneck, apply a small Transformer encoder with no positional embeddings along the document axis, and mean-pool the resulting document representations to obtain a residual $\Delta_p \in \mathbb{R}^{D}$. The final merged latent is
\[
    z_{\mathrm{merge},p}
    =
    m_p
    +
    g \cdot \mathrm{Scale}(\Delta_p, m_p),
\]
where $g$ is a learned scalar gate. The residual scaling function matches the RMS of the residual to the RMS of the mean latent at each position:
\[
    \mathrm{RMS}_D(x_p)
    =
    \sqrt{\frac{1}{D}\sum_{j=1}^{D} x_{p,j}^{2}},
    \qquad
    \mathrm{Scale}(\Delta_p,m_p)
    =
    \Delta_p
    \cdot
    \frac{\mathrm{RMS}_D(m_p)}
    {\mathrm{RMS}_D(\Delta_p)+\epsilon}.
\]
The RMS is computed only along the feature dimension $D$: we do not center the activations, use global statistics, or pool statistics across layers or latent positions. The output projection is zero-initialized and the scalar gate is initialized to $0.1$, so the merger starts from latent averaging while allowing the learned residual to modify the average during training. The merged latent is then passed through the frozen downstream hypernetwork head to generate one fixed-rank LoRA dictionary shared across all questions in the tuple.

\begin{table}[t]
\centering
\caption{Latent merger architecture hyperparameters for D2L and SHINE.}
\label{tab:latent_merger_arch}
\begin{tabular}{lcc}
\toprule
Hyperparameter & D2L & SHINE \\
\midrule
Bottleneck dimension $d_{\mathrm{lat}}$ & 128 & 256 \\
Transformer layers & 3 & 3 \\
Attention heads & 8 & 8 \\
FFN hidden dimension & 512 & 1024 \\
Activation & GELU & GELU \\
\bottomrule
\end{tabular}
\end{table}

This architecture is permutation-invariant over the input documents and does not increase the LoRA rank. It can therefore be viewed as learning a set function over document-specific hypernetwork latents while preserving the original adapter interface used by the base model.

\subsection{Experimental Setup and Baselines}
\label{subsec:merging_setup}

\paragraph{Training data and tuple construction.}
We train Latent Merger on the SQuAD training split~\citep{rajpurkar2016squad}. Both D2L and SHINE use the same merger training data. At each training step, we sample the tuple size uniformly as $N \sim \mathrm{Uniform}\{2,3,4,5\}$.
We then sample $N$ distinct passages uniformly from the SQuAD training split. For each selected passage, we sample one question uniformly from that passage's associated question list. Thus, each training tuple contains $N$ passages and one question-answer pair per passage.

\paragraph{Models and latent extraction.}
For the main experiments, we use pretrained generator checkpoints for Gemma-2-2B-it (D2L) and Qwen3-8B (SHINE). For D2L, we merge the post-ResMLP hypernetwork latents and pass the merged latent through the frozen normalization and LoRA-generation head. For SHINE, we apply the same Latent Merger architecture to the Qwen3-8B memory-state latents and pass the merged memory state through the frozen metanetwork. In both cases, the base LM, hypernetwork trunk, and LoRA-generation head remain frozen; only Latent Merger parameters are optimized. Each merger is trained for the latent representation and decoding head of its corresponding generator checkpoint.

\paragraph{Training objective.}
For SQuAD tuples, Latent Merger is trained using knowledge distillation~\citep{hinton2015distilling} with a KL-divergence objective from the corresponding per-document oracle adapters.

For each tuple, the merged adapter is shared across all $N$ questions. For question $q_i$ associated with passage $i$, the teacher is the oracle model using the document-specific LoRA generated from passage $i$. We do not mix teacher distributions for the same question. We compute teacher-forced oracle logits on the gold answer sequence. The student logits are produced by the same base model using the single merged adapter, also under teacher forcing on the gold answer sequence. The KL loss is computed only over answer-token positions:
\[
    \mathcal{L}_{\mathrm{KL}}
    =
    \frac{1}{|\mathcal{M}|}
    \sum_{(i,t)\in \mathcal{M}}
    \mathrm{KL}
    \left(
        p^{\mathrm{oracle}}_{i,t}
        \,\|\, 
        p^{\mathrm{merge}}_{i,t}
    \right),
\]
where $\mathcal{M}$ is the set of answer-token positions across the tuple. The KL temperature is $T=1.0$. For 2Wiki examples, the teacher is instead the frozen base model with both source documents in context.

\paragraph{Optimization and compute.}
For D2L, we train Latent Merger for 4000 steps. For SHINE, we train for 2000 steps. Both use AdamW~\citep{loshchilov2017decoupled} with default betas, learning rate $5\times 10^{-5}$, weight decay $0.01$, linear warmup for 100 steps, cosine learning-rate decay, gradient clipping at norm $1.0$, and batch size 2. Each batch contains two independently sampled tuples, and each tuple independently samples $N \in \{2,3,4,5\}$. The main training and evaluation experiments can be run on a single NVIDIA A100 GPU.

\paragraph{Evaluation protocol.}
We evaluate on tuples sampled from the held-out SQuAD validation set~\citep{rajpurkar2016squad} and the out-of-distribution MS MARCO validation set~\citep{bajaj2016msmarco} with $N \in \{2,5,10\}$. For each tuple, a method must produce a single merged LoRA dictionary that is used to answer every question in the tuple. The exception is the oracle, which does not merge adapters and instead answers each question using the LoRA generated from its own evidence document. We report token-overlap F1, multiplied by 100, with mean $\pm$ 95\% confidence intervals. Confidence intervals are based on 100 seeds for $N=2$, 50 seeds for $N=5$, and 20 seeds for $N=10$ on SQuAD, and 50 seeds for all $N$ values on MS MARCO.

\paragraph{Baselines and training-free LoRA merging methods.}
We compare against a suite of training-free methods that combine the document-specific adapters in a tuple without optimizing additional parameters. Let the LoRA update produced for document $i$ at a given target module be $\Delta W_i = A_i B_i^\top$,
where each adapter has rank $r$. All non-oracle methods produce a single LoRA dictionary shared across all questions in the tuple.

\begin{itemize}
    \item \textbf{Oracle:} We do not merge adapters. Instead, each question is answered using the LoRA generated from its own evidence document. This provides an empirical reference for methods that must compress all tuple information into one adapter.

    \item \textbf{Unscaled rank concatenation:} We concatenate the low-rank factors along the rank dimension, $A_{\mathrm{cat}} = [A_1,\ldots,A_N]$, $B_{\mathrm{cat}} = [B_1,\ldots,B_N]$,
    yielding a rank-$Nr$ adapter with update $\Delta W_{\mathrm{cat}} = \sum_{i=1}^{N} \Delta W_i$.
    This corresponds to the natural rank-expansion behavior of the original D2L construction and is an optimistic baseline, since it preserves all rank components rather than compressing the tuple into a fixed-rank adapter.

    \item \textbf{Mean-scaled rank concatenation:} We use the same rank-$Nr$ concatenation but divide one factor by $N$, equivalently producing $\Delta W_{\mathrm{merge}} = \frac{1}{N}\sum_{i=1}^{N} \Delta W_i$.
    This isolates the effect of rank expansion from the magnitude growth induced by unscaled concatenation.

    \item \textbf{Norm-preserving rank concatenation:} We again concatenate all rank components, but rescale the resulting update to match the average Frobenius norm of the individual document adapters:
    \[
        \Delta W_{\mathrm{merge}}
        =
        \gamma \sum_{i=1}^{N} \Delta W_i,
        \qquad
        \gamma =
        \frac{\frac{1}{N}\sum_{i=1}^{N} \|\Delta W_i\|_F}
        {\left\|\sum_{i=1}^{N} \Delta W_i\right\|_F+\epsilon}.
    \]
    This baseline preserves the full rank-expanded subspace while controlling for the scale mismatch introduced by summing multiple LoRA updates.

    \item \textbf{Factor averaging:} We average the LoRA factors directly,$A_{\mathrm{avg}} = \frac{1}{N}\sum_i A_i$, $B_{\mathrm{avg}} = \frac{1}{N}\sum_i B_i$.
    This keeps the adapter rank fixed at $r$, but averages the factorization rather than the induced weight update, and is therefore not equivalent to averaging $\Delta W_i$ in weight space.

    \item \textbf{SVD-compressed weight averaging:} We first compute the mean dense update $\bar{\Delta W} = \frac{1}{N}\sum_i \Delta W_i$,
    then take its best rank-$r$ truncated SVD approximation and refactorize it into LoRA form. This provides a strong fixed-rank weight-space baseline for compressing the average update back to the original LoRA rank.

    \item \textbf{TIES:} We adapt TIES~\citep{yadav2023ties} to the LoRA setting by applying magnitude trimming, majority-sign election, and sign-consistent merging to the aligned adapter tensors, followed by mean-scaled rank concatenation to materialize the merged LoRA dictionary. We use $d=0.5$ in the main experiments.

    \item \textbf{DARE:} We adapt DARE~\citep{yu2024language} by randomly dropping LoRA delta entries, rescaling the retained entries by the inverse keep probability, and then applying mean-scaled rank concatenation. This evaluates whether sparsifying redundant adapter deltas before merging improves robustness to cross-document interference. We experiment with $p=0.2, 0.5, 0.8$ and pick the best performance.

    \item \textbf{Latent averaging:} As a simple permutation-invariant baseline analogous to our proposed Latent Merger, we average the hypernetwork latents before LoRA generation. For D2L, we average the post-ResMLP latents and pass the result through the frozen normalization and LoRA-generation head; for SHINE, we average the memory states and pass the result through the frozen metanetwork. This is the direct training-free counterpart of our proposed Latent Merger and isolates the benefit of learning a nontrivial set function over document latents.
\end{itemize}

\subsection{Results}
\label{subsec:merging_results}

\begin{table}[t]
\centering
\caption{D2L composition results on held-out SQuAD-val and OOD MS MARCO tuples across different values of $N$. We report token-overlap F1, multiplied by 100, and mean $\pm$ 95\% CI. Concatenation-based methods increase the effective LoRA rank from $r=8$ to $Nr$, while other methods produce a single rank-$r$ LoRA.}
\label{tab:d2l_main}
\resizebox{0.99\linewidth}{!}{
\begin{tabular}{lcccccc}
\toprule
& \multicolumn{3}{c}{SQuAD} & \multicolumn{3}{c}{MS MARCO} \\
\cmidrule(lr){2-4} \cmidrule(lr){5-7}
Method & $N=2$ & $N=5$ & $N=10$ & $N=2$ & $N=5$ & $N=10$ \\
\midrule
Oracle D2L 
& \multicolumn{3}{c}{$66.8$} 
& \multicolumn{3}{c}{$42.2$}  \\
\midrule
\multicolumn{7}{l}{\textit{Rank Concatenation Methods}} \\
1/N Scaled Concat 
& 24.1 $\pm$ 2.5 & 19.3 $\pm$ 2.1 & 18.6 $\pm$ 1.4
& 20.3 $\pm$ 3.0 & 20.5 $\pm$ 1.9 & 20.7 $\pm$ 1.7 \\
Unscaled Concat 
& 30.9 $\pm$ 2.7 & 22.9 $\pm$ 2.7 & 0.6 $\pm$ 0.2
& 24.3 $\pm$ 2.8 & 6.7 $\pm$ 1.5 & 2.2 $\pm$ 0.6 \\
Norm-preserving 
& 26.6 $\pm$ 2.5 & 21.8 $\pm$ 2.2 & 20.0 $\pm$ 1.5
& 21.1 $\pm$ 2.8 & 22.1 $\pm$ 1.9 & \textbf{22.2 $\pm$ 1.8} \\
\midrule
\multicolumn{7}{l}{\textit{Weight and factor-space Methods}} \\
Factor Average 
& 41.2 $\pm$ 2.7 & 19.8 $\pm$ 2.0 & 14.8 $\pm$ 1.6
& 29.5 $\pm$ 3.5 & 22.4 $\pm$ 2.1 & 19.7 $\pm$ 1.5 \\
SVD Compress 
& 48.5 $\pm$ 2.8 & 22.0 $\pm$ 2.1 & 15.5 $\pm$ 1.7
& 31.0 $\pm$ 3.5 & 23.2 $\pm$ 2.1 & 20.1 $\pm$ 1.5 \\
TIES ($d=0.5$) 
& 43.7 $\pm$ 2.7 & 21.9 $\pm$ 2.4 & 16.4 $\pm$ 1.5
& 29.0 $\pm$ 3.6 & 21.6 $\pm$ 1.8 & 15.5 $\pm$ 1.5 \\
DARE 
& 18.7 $\pm$ 2.3 & 17.0 $\pm$ 1.9 & 17.1 $\pm$ 1.4
& 19.2 $\pm$ 2.9 & 19.2 $\pm$ 1.9 & 19.5 $\pm$ 1.7 \\
\midrule
\multicolumn{7}{l}{\textit{Latent-space Methods}} \\
Latent Average 
& 47.6 $\pm$ 2.8 & 25.2 $\pm$ 2.5 & 18.6 $\pm$ 1.7
& 31.3 $\pm$ 3.5 & \textbf{23.5 $\pm$ 2.2} & 20.7 $\pm$ 1.5 \\
Latent Merger (Ours) 
& \textbf{50.7 $\pm$ 2.8} & \textbf{29.8 $\pm$ 2.5} & \textbf{21.0 $\pm$ 1.3}
& \textbf{32.0 $\pm$ 3.7} & \textbf{23.5 $\pm$ 2.0} & \textbf{22.2 $\pm$ 1.7} \\
\bottomrule
\end{tabular}
}
\end{table}

\begin{table}[t]
\centering
\caption{SHINE composition results on held-out SQuAD-val and OOD MS MARCO tuples across different values of $N$. We report token-overlap F1, multiplied by 100, and mean $\pm$ 95\% CI.}
\label{tab:shine_main}
\resizebox{0.99\linewidth}{!}{
\begin{tabular}{lcccccc}
\toprule
& \multicolumn{3}{c}{SQuAD} & \multicolumn{3}{c}{MS MARCO} \\
\cmidrule(lr){2-4} \cmidrule(lr){5-7}
Method & $N=2$ & $N=5$ & $N=10$ & $N=2$ & $N=5$ & $N=10$ \\
\midrule
Oracle SHINE 
& \multicolumn{3}{c}{56.8} 
& \multicolumn{3}{c}{38.1} \\
\midrule
\multicolumn{7}{l}{\textit{Rank Concatenation Methods}} \\
1/N Scaled Concat 
& 44.3 $\pm$ 3.1 & 29.2 $\pm$ 3.0 & 24.7 $\pm$ 2.9
& 26.6 $\pm$ 3.9 & 22.1 $\pm$ 2.1 & 20.0 $\pm$ 1.6 \\
Unscaled Concat 
& 12.2 $\pm$ 2.0 & 0.0 $\pm$ 0.0 & 0.0 $\pm$ 0.0
& 6.4 $\pm$ 1.2 & 0.1 $\pm$ 0.1 & 0.1 $\pm$ 0.1 \\
Norm-preserving 
& 44.9 $\pm$ 3.0 & 29.1 $\pm$ 2.7 & 21.7 $\pm$ 2.3
& 28.9 $\pm$ 3.9 & 21.7 $\pm$ 2.1 & 19.4 $\pm$ 1.5 \\
\midrule
\multicolumn{7}{l}{\textit{Weight and factor-space Methods}} \\
Factor Average 
& 44.7 $\pm$ 3.1 & 29.5 $\pm$ 3.0 & 23.7 $\pm$ 2.6
& 26.7 $\pm$ 3.9 & 22.1 $\pm$ 2.0 & 20.1 $\pm$ 1.6 \\
SVD Compress 
& 44.6 $\pm$ 3.1 & 29.3 $\pm$ 3.0 & 23.6 $\pm$ 2.7
& 26.5 $\pm$ 3.7 & 22.2 $\pm$ 2.0 & 20.1 $\pm$ 1.6 \\
TIES ($d=0.5$) 
& 43.3 $\pm$ 4.6 & 26.7 $\pm$ 2.5 & 19.4 $\pm$ 2.1
& 29.0 $\pm$ 7.5 & 21.8 $\pm$ 3.2 & 19.0 $\pm$ 1.5 \\
DARE 
& 31.3 $\pm$ 4.1 & 27.5 $\pm$ 3.1 & 23.8 $\pm$ 2.8
& 31.8 $\pm$ 6.7 & 27.0 $\pm$ 3.6 & \textbf{26.0 $\pm$ 1.4} \\
\midrule
\multicolumn{7}{l}{\textit{Latent-space Methods}} \\
Latent Average 
& 45.4 $\pm$ 3.0 & 30.6 $\pm$ 2.8 & 23.5 $\pm$ 2.5
& 29.5 $\pm$ 3.6 & 22.9 $\pm$ 2.1 & 20.9 $\pm$ 1.5 \\
Latent Merger (Ours) 
& \textbf{50.2 $\pm$ 4.2} & \textbf{39.8 $\pm$ 2.8} & \textbf{30.2 $\pm$ 2.5}
& \textbf{33.5 $\pm$ 4.0} & \textbf{27.2 $\pm$ 2.2} & 25.8 $\pm$ 1.8 \\
\bottomrule
\end{tabular}
}
\end{table}

\begin{table}[t]
\centering
\caption{
Comparison of KL and CE training objectives for Latent Merger on SQuAD validation set. We report token-overlap F1, multiplied by 100, and mean $\pm$ 95\% CI.
}
\label{tab:loss_objective}
\resizebox{0.99\linewidth}{!}{
\begin{tabular}{lcccccc}
\toprule
& \multicolumn{3}{c}{D2L} & \multicolumn{3}{c}{SHINE} \\
\cmidrule(lr){2-4} \cmidrule(lr){5-7}
Loss & $N=2$ & $N=5$ & $N=10$ & $N=2$ & $N=5$ & $N=10$ \\
\midrule
KL Divergence
& \textbf{51.5 $\pm$ 4.1} & \textbf{29.8 $\pm$ 2.5} & \textbf{21.0 $\pm$ 1.3}
& \textbf{50.2 $\pm$ 4.2} & \textbf{40.6 $\pm$ 2.8} & \textbf{32.3 $\pm$ 2.6} \\
Cross-Entropy
& 48.3 $\pm$ 4.1 & 24.7 $\pm$ 2.4 & 17.9 $\pm$ 1.2
& 49.8 $\pm$ 4.6 & 39.8 $\pm$ 2.8 & 30.2 $\pm$ 2.5 \\
\bottomrule
\end{tabular}
}
\end{table}

We report the composition results for D2L and SHINE in Tables~\ref{tab:d2l_main} and~\ref{tab:shine_main}, respectively, comparing Latent Merger against rank-concatenation, weight-space, factor-space, and latent-space baselines across different composition sizes.

\paragraph{Overall performance.}
Latent Merger achieves the strongest overall composition performance across both hypernetwork settings. For D2L, it obtains the best or tied-best mean performance for every evaluated value of $N$ on both SQuAD and MS MARCO. For SHINE, it similarly leads on all SQuAD settings and on MS MARCO for $N=2$ and $N=5$, while remaining essentially competitive with the best method in the most challenging $N=10$ OOD setting. These results indicate that the learned latent-space composition rule provides a robust advantage over direct weight-, factor-, and rank-space composition strategies.

\paragraph{Generalization beyond the training regime.}
In the main experiments, Latent Merger is trained only on SQuAD tuples with $N$ sampled from $\{2,3,4,5\}$, yet the evaluation probes two forms of generalization: extrapolation to larger compositions and transfer to an out-of-domain dataset. The method remains the strongest approach when extrapolated to $N=10$ on SQuAD for both D2L and SHINE, suggesting that the learned merger does not merely memorize the training cardinalities. It also transfers well to MS MARCO, consistently matching or exceeding competing approaches except in the most OOD SHINE setting, where its performance is nearly indistinguishable from the best baseline. This suggests that latent composition captures structure that is not specific to the SQuAD training distribution.

\paragraph{Limitations of rank concatenation.}
Rank concatenation provides a larger effective LoRA rank than the other composition methods, and might therefore be expected to preserve more task-specific information. Surprisingly, this additional capacity does not translate into reliable performance. This is especially notable for D2L, where the training procedure already encourages concatenation-like composition by chunking documents and composing the resulting LoRAs. Across both D2L and SHINE, however, none of the unscaled, $1/N$-scaled, or norm-preserving concatenation variants is consistently competitive across all values of $N$. The results suggest that simply increasing rank does not solve the composition problem: unscaled concatenation becomes unstable as more LoRAs are added, whereas aggressive scaling weakens the contribution of each individual LoRA.

\paragraph{Effectiveness of latent-space composition.}
The comparison between latent-space methods and weight- or factor-space baselines supports the central hypothesis that hypernetwork-generated LoRAs are easier to compose before they are decoded into LoRA factors. Even the training-free Latent Average baseline is highly competitive and often outperforms direct factor averaging, SVD compression, TIES, and DARE. Latent Merger further improves upon this by learning a composition function in the hypernetwork latent space, yielding the most consistent gains across datasets, composition sizes, and hypernetwork backbones. This pattern indicates that the latent representation provides a smoother and more semantically aligned space for merging than the induced LoRA weight space itself.

\paragraph{Multi-hop composition.}
Table~\ref{tab:multi_hop_2wiki} reports D2L results on 2WikiMultihopQA~\citep{ho2020constructing} compositional questions. Training the merger on an XSum+2Wiki+MMLU mixture with an in-context teacher that reads both source documents for 2Wiki examples increases judge accuracy from $36.0\%$ to $42.0\%$, an absolute gain of $6.0$ percentage points. The SQuAD-trained merger does not improve over latent averaging in this evaluation.

\begin{table}[t]
  \centering
  \small
  \caption{D2L composition on 2WikiMultihopQA compositional questions. Judge accuracy is reported as a percentage. The mixed-task-trained merger uses the frozen base language model with both source documents in context as its teacher for 2Wiki examples; the SQuAD-trained merger uses per-document teachers.}
  \label{tab:multi_hop_2wiki}
  \begin{tabular}{llc}
    \toprule
    Method & Merger training data & Judge accuracy (\%) \\
    \midrule
    Latent averaging & None & $36.0$ \\
    Latent Merger & SQuAD & $36.0$ \\
    Latent Merger & XSum+2Wiki+MMLU & $42.0$ \\
    \bottomrule
  \end{tabular}
\end{table}

\paragraph{KL vs. Cross-Entropy training objective for Latent Merger:} In addition to the knowledge distillation KL-divergence objective that we used to train Latent Merger, using the cross-entropy loss on the ground truth answers as in standard supervised fine-tuning (SFT) is also a natural choice. We compare KL divergence and cross-entropy as the training objective for Latent Merger training in Table~\ref{tab:loss_objective}. We observe that the KL loss consistently outperforms the CE loss across all values of $N$ on both D2L and SHINE. The results suggest that matching the full output distribution provides a more reliable training signal than optimizing only the ground-truth answer likelihood. Overall, these results indicate that KL loss is the preferable objective for training Latent Merger.

\paragraph{Additional evaluations.}
Appendix~\ref{app:additional_backbones} reports evaluations on additional D2L backbones. Appendix~\ref{app:capacity} extends fixed-rank evaluation to $N=20,50$. Appendix~\ref{app:composition_profiling} reports one-time composition latency and transient GPU memory. At $N=20$, the complete latent path takes $0.77$\,ms for D2L and $44.5$\,ms for SHINE on an NVIDIA H200. The latency and peak-memory trade-offs depend on the architecture and baseline.

\section{Discussion and Conclusion}
\label{sec:discussion}
In this work, we study the geometry and composition of hypernetwork-generated LoRA adapters. Across two representative systems, D2L and SHINE, we find that generated document adapters differ sharply from LoRAs obtained by direct fine-tuning. While NTP-FT and QA-SFT adapters are diffuse and nearly orthogonal across documents, hypernetwork-generated adapters are much more spectrally concentrated and occupy compact regions of adapter space. These results suggest that hypernetwork-generated LoRAs should not be treated simply as ordinary low-rank weight updates, but as decoded points on a structured adapter manifold induced by the shared generator.

This structural difference directly motivates our merging method. Instead of composing adapters after they have been decoded into LoRA weights or factors, we merge their intermediate hypernetwork latents and then decode the merged representation through the frozen LoRA-generation head. Both latent averaging and Latent Merger outperform standard weight-space and factor-space baselines across D2L and SHINE, while producing a single fixed-rank adapter. The learned merger further improves over latent averaging, showing that a lightweight permutation-invariant set function can capture useful cross-document corrections without modifying the base LLM or the original hypernetwork. These findings suggest that latent-space composition can be a promising mechanism for bounded accumulation of hypernetwork-generated document adapters.

\paragraph{Limitations and future work.}
Our reported composition evaluations cover QA, while non-QA generalization and resolution of conflicting sources remain untested. Our experiments primarily evaluate composing multiple LoRAs into a single fixed-rank LoRA adapter, which can become a capacity bottleneck as more documents are merged. A natural next step is to combine latent merging with adaptive rank allocation, retrieval over merged adapter banks, and online memory-management policies that decide when to merge, retain, or discard document adapters in a streaming setting.

\section*{Acknowledgments}
We thank Kyunghyun Cho for helpful discussion. This research was supported by Visko Platform, Pionex, a Google TPU Award, the NYU-KAIST Award A25-0081-002, and the Institute of Information \& Communications Technology Planning \& Evaluation (IITP) with a grant funded by the Ministry of Science and ICT (MSIT) of the Republic of Korea in connection with the Global AI Frontier Lab International Collaborative Research Agreement (No. RS-2024-00469482 \& RS-2024-00509279). This work was also supported by the Institute for Information \& communications Technology Planning \& Evaluation (IITP) grant funded by the Korea government (MSIT) (RS-2019-II190075, Artificial Intelligence Graduate School Program (KAIST)). The compute is supported by the NYU High Performance Computing resources, services, and \mbox{staff expertise.}

\section*{Funding Transparency Statement}
\textit{Funding in direct support of this work:} Visko Platform, Pionex, a Google TPU Award, the NYU-KAIST Award A25-0081-002, and grants from
IITP and MSIT of the Republic of Korea through the Global AI Frontier Lab (RS-2024-00469482, RS-2024-00509279) and the Artificial Intelligence Graduate School Program (KAIST) (RS-2019-II190075), and NYU HPC.

\textit{Additional revenues:} Student support from the Meta AI Mentorship Program, Azure credits from the Microsoft Accelerating Foundation Models Research program, Google Cloud credits, Anthropic API credits, and compute from the Center for AI Safety.

\bibliography{ref}
\bibliographystyle{apalike}

\clearpage
\appendix

\section{Additional backbone evaluation}
\label{app:additional_backbones}

We additionally train Latent Merger using released D2L checkpoints for Mistral-7B~\citep{jiang2023mistral7b} and Qwen3-4B~\citep{yang2025qwen3}, keeping the underlying language models and hypernetworks frozen. Table~\ref{tab:additional_backbones} reports the SQuAD-2067 composition evaluation. Latent Merger achieves higher mean scores than latent averaging at all three composition sizes on both backbones. The mergers use 2,000 training steps. Evaluation uses the 2,067-passage, 10,570-question SQuAD dataset, one random tuple per seed per $N$, and 64-token greedy decoding.

\begin{table}[htbp]
  \centering
  \small
  \setlength{\tabcolsep}{7pt}
  \caption{Additional D2L backbones on SQuAD-2067. We report token-overlap F1, multiplied by 100, and mean $\pm$ 95\% CI.}
  \label{tab:additional_backbones}
  \begin{tabular}{lccc}
    \toprule
    Method & $N=2$ & $N=5$ & $N=10$ \\
    \midrule
    \multicolumn{4}{c}{\textbf{D2L Mistral-7B}} \\
    \midrule
    Per-document oracle & $40.5\pm5.8$ & $38.7\pm3.7$ & $38.3\pm2.0$ \\
    Latent Merger & $29.1\pm4.4$ & $20.6\pm3.5$ & $15.5\pm1.8$ \\
    Latent averaging & $22.3\pm4.9$ & $13.8\pm2.6$ & $10.8\pm1.6$ \\
    SVD-compressed weight averaging & $22.1\pm5.2$ & $13.0\pm2.8$ & $10.8\pm1.5$ \\
    Factor averaging & $20.3\pm4.4$ & $11.8\pm2.8$ & $10.3\pm1.5$ \\
    TIES ($d=0.5$) & $16.2\pm3.7$ & $9.5\pm2.5$ & $8.3\pm1.2$ \\
    \midrule
    \multicolumn{4}{c}{\textbf{D2L Qwen3-4B}} \\
    \midrule
    Per-document oracle & $25.4\pm3.8$ & $26.4\pm2.4$ & $26.5\pm1.3$ \\
    Latent Merger & $23.0\pm3.5$ & $16.0\pm3.0$ & $12.6\pm1.4$ \\
    Latent averaging & $12.9\pm2.7$ & $8.7\pm1.2$ & $7.0\pm0.7$ \\
    SVD-compressed weight averaging & $15.3\pm3.4$ & $7.5\pm1.4$ & $6.4\pm0.6$ \\
    Factor averaging & $10.7\pm2.2$ & $6.6\pm1.2$ & $6.1\pm0.6$ \\
    TIES ($d=0.5$) & $11.1\pm3.0$ & $6.6\pm1.3$ & $7.2\pm0.9$ \\
    \bottomrule
  \end{tabular}
\end{table}

\section{Fixed-rank composition at larger cardinalities}
\label{app:capacity}

Table~\ref{tab:extended_fixed_rank} reports the extended SQuAD-2067 evaluation at $N=20$ and $N=50$, alongside closed-book and per-document-oracle references. The flat merger remains above the closed-book score, but its score decreases with cardinality and remains substantially below the per-document oracle.

\begin{table}[htbp]
  \centering
  \small
  \caption{Extended fixed-rank composition on SQuAD-2067, in token-overlap F1 multiplied by 100.}
  \label{tab:extended_fixed_rank}
  \begin{tabular}{lcc}
    \toprule
    Setting & D2L Gemma-2B & SHINE Qwen3-8B \\
    \midrule
    Closed-book & $10.6$ & $5.0$ \\
    Per-document oracle & $68.0$ & $54.8$ \\
    Flat Latent Merger, $N=20$ & $15.1$ & $24.2$ \\
    Flat Latent Merger, $N=50$ & $13.0$ & $21.5$ \\
    \bottomrule
  \end{tabular}
\end{table}

\section{Multi-hop QA extension}
The 2WikiMultihopQA extension uses compositional questions with evidence from two source documents. Instead of using a per-document oracle teacher, we train the merger against the frozen base language model with both documents in context for 2Wiki examples. The student answers using the merged adapter without access to those source documents or their titles.Table~\ref{tab:multi_hop_2wiki} compares this mixed-task-trained merger with latent averaging and the SQuAD-trained merger. Evaluation uses 300 dev-split questions filtered to exactly two gold supporting paragraphs, with 32-token greedy decoding.

\section{Latency Measurements}
\label{app:composition_profiling}

We profile the latency of different LoRA merging methods on an NVIDIA H200 GPU. Results are presented in Tables~\ref{tab:profiling_d2l_latency} and ~\ref{tab:profiling_shine_latency}. At $N=20$, the complete latent path is faster than the measured TIES, SVD-compression, and DARE implementations for both architectures. 

\begin{table}[htbp]
  \centering
  \small
  \caption{D2L Gemma-2B composition latency in milliseconds on an NVIDIA H200.}
  \label{tab:profiling_d2l_latency}
  \begin{tabular}{lrrrr}
    \toprule
    Operation & $N=2$ & $N=5$ & $N=10$ & $N=20$ \\
    \midrule
    Composition block & $0.54$ & $0.54$ & $0.57$ & $0.50$ \\
    Generator head & $0.17$ & $0.17$ & $0.16$ & $0.17$ \\
    Complete latent path & $0.83$ & $0.83$ & $0.86$ & $0.77$ \\
    \midrule
    Elementwise mean & $0.03$ & $0.03$ & $0.04$ & $0.05$ \\
    Rank concatenation & $0.04$ & $0.06$ & $0.09$ & $0.16$ \\
    TIES & $12.5$ & $13.2$ & $14.2$ & $16.3$ \\
    SVD compression & $17.4$ & $50.6$ & $96.8$ & $209.1$ \\
    DARE & $17.6$ & $49.1$ & $88.3$ & $180.6$ \\
    \bottomrule
  \end{tabular}
\end{table}

\begin{table}[htbp]
  \centering
  \small
  \caption{SHINE Qwen3-8B composition latency in milliseconds on an NVIDIA H200.}
  \label{tab:profiling_shine_latency}
  \begin{tabular}{lrrrr}
    \toprule
    Operation & $N=2$ & $N=5$ & $N=10$ & $N=20$ \\
    \midrule
    Composition block & $4.5$ & $5.6$ & $7.2$ & $10.4$ \\
    Generator head & $34.0$ & $33.8$ & $33.9$ & $33.8$ \\
    Complete latent path & $39.0$ & $39.9$ & $41.1$ & $44.5$ \\
    \midrule
    Elementwise mean & $7.3$ & $13.4$ & $23.4$ & $43.1$ \\
    Rank concatenation & $4.3$ & $4.8$ & $5.5$ & $7.1$ \\
    TIES & $175.7$ & $437.4$ & $873.1$ & $1745.3$ \\
    SVD compression (rank $r$) & $163.3$ & $463.8$ & $840.3$ & $1726.4$ \\
    DARE ($p=0.5$, rescaled) & $28.9$ & $66.1$ & $128.0$ & $249.3$ \\
    \bottomrule
  \end{tabular}
\end{table}

\section{Broader impacts}
\label{app:broader_impacts}
We study how to combine document knowledge in hypernetwork-generated adapters while keeping the base language model fixed. Composing several document adapters into one fixed-rank adapter could make document-based language systems more accessible under limited storage budgets. However, adapters may encode inaccurate, biased, or sensitive content from their source documents, and our results show capacity limits as more documents are combined. Merging document representations may also make source attribution and selective removal of information more difficult, although we do not evaluate these properties.

\end{document}